\pdfoutput=1

\documentclass[11pt]{article}

\usepackage{EMNLP2023}

\usepackage{times}
\usepackage{latexsym}
\usepackage{comment}

\usepackage{amsmath}
\usepackage{amssymb}
\usepackage{mathtools}
\usepackage{amsthm}
\usepackage{multirow}
\usepackage{hyperref}
\usepackage{bbding}
\usepackage{algorithmic}
\usepackage{algorithm}

\usepackage{microtype}
\usepackage{graphicx}
\usepackage{subfigure}
\usepackage{booktabs} %
\usepackage{xcolor}
\usepackage{enumitem}

\usepackage{hyperref}
\usepackage{bbding}

\usepackage[capitalize,noabbrev]{cleveref}
\usepackage[export]{adjustbox}

\usepackage[T1]{fontenc}

\usepackage[utf8]{inputenc}

\usepackage{microtype}

\usepackage{inconsolata}

\title{Sparse Universal Transformer}

\author{Shawn Tan$^\text{1~*}$\\
  \texttt{tanjings@mila.quebec} \\
  \And
  Yikang Shen$^\text{2~*}$ \\
  \texttt{yikang.shen@ibm.com}  \\
  \AND
    Zhenfang Chen$^\text{2}$ \\
  \texttt{zfchen@ibm.com}  \\
  \And
  Aaron Courville$^\text{1}$ \\
  \texttt{courvila@iro.umontreal.ca} \\
  \And
  Chuang Gan$^\text{2}$ \\
  \texttt{chuangg@ibm.com} \\
  \AND
  ~ \\
  $^\text{1}$Mila, University of Montreal \\ 
  \And
  ~\\
  $^\text{2}$MIT-IBM Watson AI Lab
  }

\begin{document}
\maketitle

\begin{abstract}
The Universal Transformer (UT) is a variant of the Transformer that shares parameters across its layers.
Empirical evidence shows that UTs have better compositional generalization than Vanilla Transformers (VTs) in formal language tasks.
The parameter-sharing also affords it better parameter efficiency than  VTs.
Despite its many advantages, scaling UT parameters is much more compute and memory intensive than scaling up a VT.
This paper proposes the Sparse Universal Transformer (SUT), which leverages Sparse Mixture of Experts (SMoE) and a new stick-breaking-based dynamic halting mechanism to reduce UT's computation complexity while retaining its parameter efficiency and generalization ability. 
Experiments show that SUT achieves the same performance as strong baseline models while only using half computation and parameters on WMT'14 and strong generalization results on formal language tasks (Logical inference and CFQ).
The new halting mechanism also enables around 50\% reduction in computation during inference with very little performance decrease on formal language tasks.

\end{abstract}

\section{Introduction}

Recent theoretical work has pointed out that finite-depth Transformers have an issue of expressibility that will result in failure to generalize~\citep{hahn2020theoretical,hao2022formal,merrill2022saturated,liu2022transformers}.
\citet{deletang2022neural} ran several neural architectures on a suite of different synthetic languages generated from different levels of the Chomsky hierarchy and empirically confirmed these results, showing that VTs have difficulty generalizing to Regular languages.
Universal Transformers (UTs; \citealt{dehghani2018universal}) are Transformers that share parameters at every layer of the architecture.
\citet{csordas2021devil}  performed several compositional generalization experiments on VTs and UTs along with absolute and relative position embeddings, and showed that UTs with relative positional embeddings performed better on these tasks.

\begin{figure}[t]
\begin{center}
\includegraphics[width=.85\linewidth]{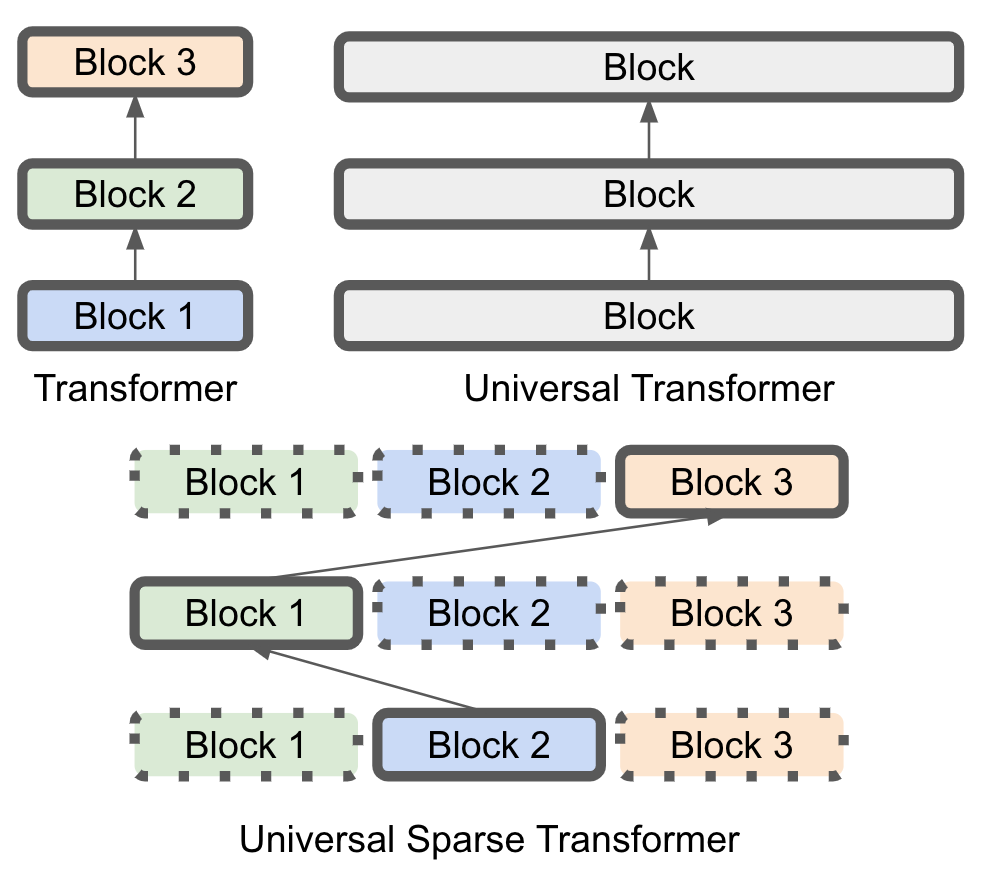}
\caption{%
A VT has separate Transformer blocks for each layer, with different parameters.
For a UT with the same number of parameters, the UT block will be $\sim$3 times the dimensions of each VT block. Running this block for 3 layers would then incur approximately 9 times the runtime memory. 
Using SMoEs can recover approximately the same computational cost as the VT. 
}
\label{fig:ut_to_sut}
\end{center}
\end{figure}
However, the task of scaling UTs is challenging due to its computation complexity~\citep{kaplan2020scaling,tay2022scaling,takase2021lessons}.
Consider a VT with $P$ parameters for each layer and $L$ layers.
Evaluating such a VT has computation complexity associated with the model size $LP$.
A size-equivalent UT would have a UT block with $LP$ parameters and computation complexity of approximately $LP$ to run the block once.
To run such a UT for equivalent $L$ layers would incur a complexity of $L^2P$.
This increased computation complexity directly translates to increased training and inference time.
According to \citet{takase2021lessons}, UT requires two times the training time and far more GPU memory than VT in WMT English-German translation task.

Sparsely activated neural networks were introduced to reduce the computation complexity of large models.
These networks activate parts of the network conditioned on the input, computing only parts of the model, thereby disentangling the number of parameters from the computation complexity.
This method allows for drastically increasing the number of parameters without proportionally increasing the computation complexity.
\citet{shazeer2017outrageously} introduced Sparse Mixture of Experts (SMoE), using the top-$k$ operator to allow for sparse computation of experts.
This allows for replacing the FeedForword (FFD) layer in the Transformer with an ensemble of $E_\text{ffd}$ FFDs, but only $k$ FFDs (where $k < E$) would have to be evaluated, conditioned on the input.
\citet{zhang2022mixture} then introduced the Mixture of Attention heads (MoA), which allows Transformers to replace its Multihead Attention (MHA) layer with an ensemble of $E_\text{att}$ attention heads and only activates $k$ heads condition on the input, further sparsifying the model.

This paper introduces the Sparse Universal Transformer (SUT), which applies the above sparse computation techniques to UT. 
Additionally, we replace the per-position halting mechanism in UT with a new stick-breaking formulation that has a probabilistic interpretation, allowing us to introduce an Adaptive Computation Time (ACT; \citealt{graves2016adaptive}) penalty to minimize layer use.
It also provides an easy way to adjust the trade-off between the amount of computation and model performance during inference, further improving the efficiency of the SUT at inference time.

To demonstrate effective scaling, we perform experiments on WMT'14 English to German translation, showing that an SUT can achieve better performance for the same parameter count,  while incurring less computation cost than an equivalent dense UT.
Since the UT setting is a specific case of SUT, we show on the Compositional Freebase Questions (CFQ; \citealt{keysers2019measuring}) tasks that UTs have better compositional generalization properties, improving upon CFQ results from \citet{csordas2021devil}.
Using the Logical Inference task \citep{bowman2015tree}, we analyse the behaviour of our UT on length and compositional generalization.
Finally, we show that the halting mechanism can be used to achieve further efficiency during inference time, and study the trade-off between efficiency and performance.

\section{Background \& Related Work}

\begin{figure}[t]
\begin{center}
\includegraphics[width=1\linewidth]{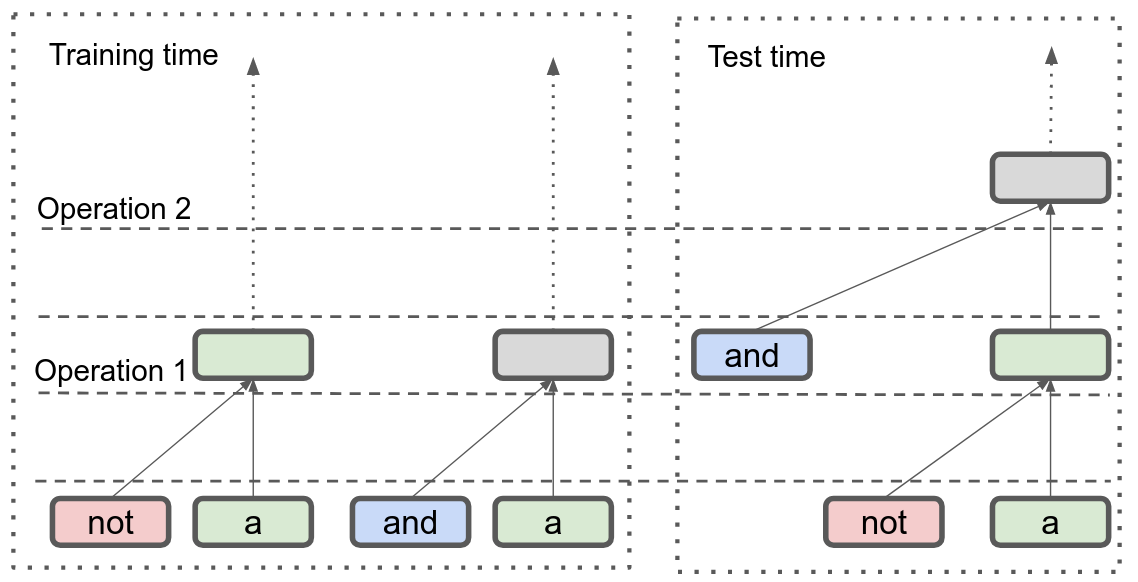}
\caption{Example of the compositional generalization splits from \cite{shen2019ordered}. 
The combination of \texttt{not} and \texttt{and} are never seen in successive combination during training, and a VT may learn a shortcut that prevents generalisation during test.}
\label{fig:invariance}
\end{center}
\vskip -0.2in
\end{figure}
\paragraph{Overcoming VT limitations with UT}
\citet{dziri2023faith} and \citet{liu2022transformers} find that Vanilla Transformers learn shortcuts for tasks that require multi-step compositional operations, and fail to generalize on larger instances of the problem that require more steps.
Theoretical results have also shown that Vanilla Transformers have limitations in what they can compute that support these findings \citep{hahn2020theoretical,hao2022formal,merrill2022saturated}.
Universal Transformers \citep{dehghani2018universal} are Transformers with tied weights across all layers, and an additional halting mechanism to decide when to stop. 
In an ideal scenario of infinite layers (now that all layers have the same parameters) the UT, like the Neural GPU \citep{kaiser2015neural}, is Turing-complete, which overcomes many of the abovementioned issues.

In practice, even with limited depth, UTs have exhibited properties that afford them better performance in compositional generalization tasks \citep{csordas2021devil}.
UTs allow operations learned in the Transformer during training to be depth-order invariant.
If some operations during training are learned to be performed in a certain order, during test time, the UT could generalize to an unseen order of operations.

\paragraph{Challenges with Scaling the UT} 
Despite these compositional abilities, performance tends to decrease on real-world tasks when using UTs. 
ALBERT \cite{lan2019albert} improved parameter efficiency by sharing parameters across layers.
This was motivated by an observation that Transformers tend to learn to perform similar operations in the layers, and that sharing these parameters would reduce this redundancy\footnote{\url{https://ai.googleblog.com/2019/12/albert-lite-bert-for-self-supervised.html}}.
However, the authors observe a dip in performance when sharing parameters, contrary to \citet{dehghani2018universal}.

Could the issue be one of model capacity?
Experiments with ALBERT show that scaling up ALBERT can outperform the BERT baseline, even on real-world tasks \citep{lan2019albert}.
\citet{kaplan2020scaling} also show that a shared-parameter Transformer has better scaling properties in terms of parameter-to-performance, but poorer properties in terms of computation-to-performance, since parameter count causes the computation to increase.
\citet{tay2022scaling} scale up different sequence models, and remark on difficulties with scaling up UTs, limiting the experiments they can perform on UT.
\citet{takase2021lessons} outline several strategies of scaling up shared-parameter Transformers to deal with these issues by using different parameter-sharing schemes. 

Our experiments show that SMoE techniques can be applied successfully to the UT to scale it up, achieving the UT's parameter efficiency while not incurring the same computation costs.
We also perform experiments that support the compositional generalization claims of prior work, and provide better baselines for those tasks.

\section{Method}
\begin{figure*}
\begin{center}
\includegraphics[width=.5\columnwidth]{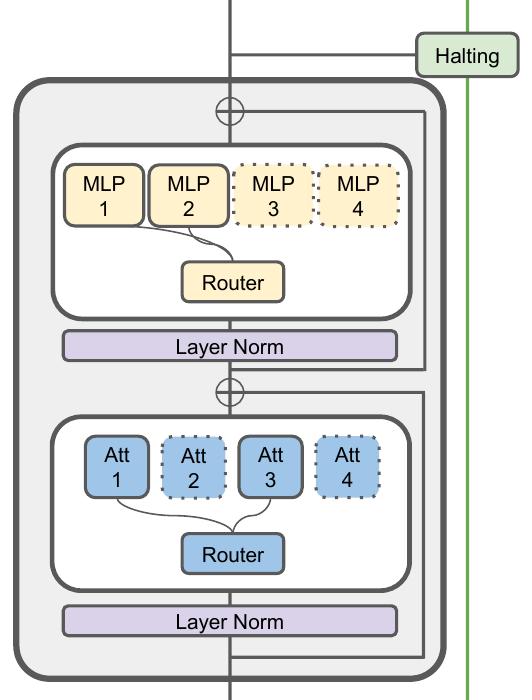}
\hspace{4em}
\includegraphics[width=.8\columnwidth]{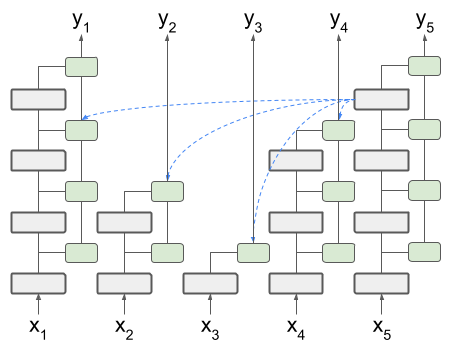}

\caption{\emph{Left}: Schematic of a SUT block. \quad \emph{Right}: While the input of each SUT block is the output of the previous layer, the attention mechanism attends to the halted state of the timestep. When the halting probability exceeds $\alpha_\text{thresh}$, the hidden state is simply copied. Finally, the halted state is used as the output of the SUT.}
\label{fig:sut_block_halting}
\end{center}
\vskip -0.2in
\end{figure*}

Like UT, we reuse the same SUT block for every layer of the Transformer.
Within each SUT block, we use SMoEs to achieve sparsity for the feed-forward network (FFD) and attention heads separately.
We use the Mutual Information Maximization loss proposed in \citet{chen2022mod} and modified for unsupervised tasks in \citet{shen2023moduleformer}.
Finally, we propose a stick-breaking process formulation of dynamic halting, which affects how the attention mechanism works in the SUT, and the Adaptive Computation Time (ACT) auxiliary loss we use to minimize the number of layers used.

\subsection{Sparse Mixture of Experts}
\newcommand{\h}{\mathbf{h}}
\newcommand{\Ss}{\mathbf{S}}
\newcommand{\q}{\mathbf{q}}
\newcommand{\K}{\mathbf{K}}
\newcommand{\V}{\mathbf{V}}
\newcommand{\W}{\mathbf{W}}
\newcommand{\Pos}{\mathbf{P}}
\newcommand{\att}{{\boldsymbol \alpha}}
\newcommand{\ctx}{\mathbf{c}}
\newcommand{\out}{\mathbf{o}}
\newcommand{\y}{\mathbf{y}}
A Mixture of Experts module consists of $E$ sub-modules $f_1,\hdots,f_E$.
There is also a gating network, which we will denote by $g(e \mid \h)$ -- for any input $\h$ to the MoE module, the gating network would predict a distribution over the $E$ experts.
When $k < E$, we refer to this as a Sparse Mixture of Experts (SMoE),  and $g(e \mid \h) > 0$ for only $k$ experts, while maintaining that $\sum_e^E g(e \mid \h) = 1$.
The final output of the SMoE is then given by $y = \sum_{e=1}^E g(e \mid \h) \cdot f_e(\h)$, where $g(e \mid \h) = 0$,  $f_e(\h)$ will not need to be evaluated, reducing computation cost during training and inference.
We replace the Feed-forward layer (FFD) in the Transformer block with a mixture of FFDs. Each Mixture of FFD can be described with a 3-tuple, $(E, k, D)$:
$E$ experts, $k$ for the number of experts to be used in the top-$k$ operation, and $D$ for the dimension of the hidden layer for each FFD expert.
For the attention layer, we use the Mixture of Multi-head Attention (MoMHA) proposed by \citet{zhang2022mixture} and \citet{shen2023moduleformer}.
Each MoMHA module can be described by a 5-tuple, $(E, k, H, D, W)$,
with $E$ representing the number of experts, $K$ representing the parameter $k$ in a top-$k$ operation, $H$ representing the number of attention heads per expert, and $D$ for the dimensions per head. 
Like in MoA, MoMHA maintains only a single set of $H$ key-value projections shared among the experts, while there are $E$ query projections of $H$ heads each.
$W$ represents the relative position embedding window size, parameterizing $2W+1$ embeddings for $W$ positions before and after the present position. 
Figure \ref{fig:sut_block_halting} (\emph{Left}) shows the schematic of a SUT block.

This technique has been used to reduce computation costs both during training and inference time for large models. 

\paragraph{Mutual Information Maximisation}

Like other models that rely on conditional activation, auxiliary losses are needed in order to aid learning a module that decides which experts are activated, and to ensure that all experts are used, balancing the load for processing.
For this, we use  the Mutual Information Maximization introduced in \citet{chen2022mod} for the auxiliary loss (to be maximised):
\begin{align}
\mathcal{L}_\text{MIM} =& \underbrace{\sum_{e=1}^E  g(e) \log g(e)}_{-H(e)} \nonumber \\
 &\underbrace{- \frac{1}{|\mathcal{X}|} \sum_{\h \in \mathcal{X}} \sum_{e=1}^E  g(e\mid \h) \log g(e\mid \h)}_{H(e\mid\h)},
\end{align}
where, 
$$g(e) = \frac{1}{|\mathcal{X}|} \sum_{\mathbf{h} \in \mathcal{X}} g(e|\mathbf{h})$$.
Specifically, we use the unsupervised version proposed by \citet{shen2023moduleformer} that assumes a uniform distribution over all tokens and layers, resulting in the following auxiliary objective.
In the SUT setting, the gating network is used $|\mathcal{X}| = L \cdot T$ times, where $L$ is the number of layers, and $T$ is the number of timesteps.

Intuitively, the entropy term increases the entropy of the marginal probability of the gating network predictions, which at its maximum means that the weight for each gating network across the entire minibatch is uniform.
The conditional entropy term decreases the conditional entropy, which causes the prediction of the gating network to be sharp, and also penalizes the uniform distribution solution for the gating network.
\subsection{Stick-breaking Dynamic Halting}
\begin{algorithm}[t]
   \small
   \caption{Halting mechanism at a given timestep $t$}
   \label{alg:dynhalting}
\begin{algorithmic}
   \FOR{$l=1$ {\bfseries to} $L$}
   \IF{$\sum_{l'=1}^{l-1}  \alpha^{(t)}_{l'} < \alpha_{\textrm{thresh}}$}
   \STATE $\displaystyle \hat{\alpha}^{(t)}_{l-1} = \mathrm{halt}(\mathbf{h}^{(t)}_{l-1})$
   \STATE $\displaystyle \alpha^{(t)}_{l-1} = \hat{\alpha}^{(t)}_{l-1} \prod_{l'=1}^{l-2} (1 - \hat{\alpha}^{(t)}_{l'})$
   \STATE $\displaystyle \mathbf{a}^{(t)}_l = \mathrm{Attention}(
        \underbrace{\mathbf{h}^{(t)}_{l-1}}_{Q}, 
        \underbrace{\mathbf{S}_{l-1}}_{K},
        \underbrace{\mathbf{S}_{l-1}}_{V}) $
   \STATE $\displaystyle \mathbf{h}^{(t)}_l = \mathrm{FeedForward}(\mathbf{h}^{(t)}_{l-1}, \mathbf{a}^{(t)}_l)$
   \STATE $\mathbf{s}^{(t)}_l = \left(1 - \sum_{l'=1}^{l-1}  \alpha^{(t)}_{l'}\right) \cdot \mathbf{h}^{(t)}_l$
   \STATE $ \qquad\qquad\qquad\quad +\left(\sum_{l'=1}^{l-1}  \alpha^{(t)}_{l'} \cdot \mathbf{\mathbf{h}}^{(t)}_{l'} \right)$
   \ELSE 
      \STATE $\displaystyle \mathbf{h}^{(t)}_l =  \mathbf{h}^{(t)}_{l-1}$
      \STATE $\displaystyle \mathbf{s}^{(t)}_l =  \mathbf{s}^{(t)}_{l-1}$
   \ENDIF
   \ENDFOR
\end{algorithmic}
\end{algorithm}

There have been several methods for imbuing models with the ability to make a prediction without having to use all layers of the model \cite{graves2016adaptive,tan2016towards,dehghani2018universal,elbayad2019depth,schuster2022confident}.
Motivations for this include: (1) different inputs require different amounts of iteration to make a prediction, (2) reducing computation cost. 

UT implements a similar mechanism, but the UT version of halting is difficult to interpret.
Here we choose a principled version of the dynamic halting mechanism based on the stick-breaking process, viewing it as a probability distribution.
First, $ \hat{\alpha}^{(t)}_{l}$ are the halting probabilities predicted by $\mathrm{halt}(\mathbf{h}^{(t)}_{l})$ , a function which is implemented by an MLP that takes in the previous layer's embedding.
Then, the probability of any layer halting is computed by 
\begin{align}
\alpha^{(t)}_{l} = \hat{\alpha}^{(t)}_{l} \prod_{l'=1}^{l-1} (1 - \hat{\alpha}^{(t)}_{l'}).
\end{align} 
A similar formulation is described in \citet{graves2016adaptive} and \citet{tan2016towards}.
Algorithm \ref{alg:dynhalting} shows how the mechanism is implemented at any given timestep.
$\mathbf{h}_{l-1}$ is the output of the previous layer for the current timestep.

Conditioned on the fact that we are computing $\mathbf{h}^{(t)}_{l}$, time-step $t$ must not have halted before or at $l-1$.
So we can use $\mathbf{h}^{(t)}_{l}$, the unhalted state, as input to the computation of the attention query of the block.
However, since time-step $t$ can attend to all other timesteps, and it these other steps may have halted, we use the halted states $\mathbf{S}_{l-1}$ for the previous layers.

However, because the halting is a `soft' decision, we can relax the requirement for evaluating all possible halted states and use the expected halted state as a substitute.
Previous halting mechanisms use a `gating' mechanism of convex sums between previously gated outputs and the current step's output $\h_l = \alpha_l\cdot \hat{\h}_l + (1 - \alpha_l) \cdot \h_{l-1} $ \citep{dehghani2018universal}.
This can lead to vanishingly small gradients going up the layers as $(1 - \alpha_l)$ multiplies.
We can instead compute the expected halted embedding at any $l$, 
\begin{align}
\mathbf{s}^{(t)}_l = \underbrace{\left(1 - \sum_{l'=1}^{l-1}  \alpha^{(t)}_{l'}\right) \cdot \h^{(t)}_l}_\text{previous layer if not halted}  + 
\underbrace{\sum_{l'=1}^{l-1}  \alpha^{(t)}_{l'} ~\h^{(t)}_{l'} }_\text{halted at $< l$}
\end{align}
If $ \alpha^{(t)}_{l} = 1$  for some $l$, $\mathbf{s}^{(t)}_l = \h^{(t)}_{l}$, recovering the behavior of the discrete halting decision.
We use  $\mathbf{s}^{(t)}_l$ as input to the attention key and value transformations. 

This probabilistic interpretation also allows us to impose a loss on the expected number of layers used at each step, biasing the model towards fewer iterations, thereby saving computational cost.
\begin{align}
\mathcal{L}_\text{ACT} = \frac{1}{T}\sum_{t=1}^{T} \sum_{l=1}^{L}  \alpha^{(t)}_{l} \cdot l.
\end{align}

We use a threshold $\alpha_{\text{thresh}} = 0.999$, such that the cumulative sum of the halting probabilities has exceeded this, no computation will be performed for that time step, and the previous layer's embeddings will be copied.
Due to the routing operation required in the implementation fo SMoEs, we can simply route halted states to a ``No Op'' expert, leading to real savings in computation cost when halting hits the threshold early.
We find that adjusting this threshold \emph{after} training can maintain performance while saving computation steps. 

\begin{table}[t]
    \caption{BLEU score on WMT14 En-De translation datasets. 
    MACs (Multiply–Accumulate Operations)\footnotemark[1] measures the computational complexity of each model.
    $^a$\citet{vaswani2017attention}, $^b$\citet{liu2020very}, $^c$\citet{peng2020mixture}, $^d$\citet{zhang2022mixture},
    $^e$\citet{dehghani2018universal}, $^f$\citet{myle2018scaling}, $^g$\citet{wu2018pay}
    }
    \centering
    \small
    \vskip -0.1in
    \begin{tabular}{rccr}
    \toprule
        Model & \#Params & BLEU & MACs\footnotemark[1]  \\ 
        \midrule
        Transformer base$^a$ & 65M & 27.3 & 604M \\
        Admin 6L-6L$^b$ & 61M & 27.7 & 604M \\
        MAE-7 $^c$ & 63M & 28.4 & - \\
        MoA base$^d$ & 65M & 28.4 & 628M \\ 
        UT$^e$ & 65M & 28.9 & - \\
        \midrule
        UT base + SB halting & 64M & 29.3 &  1998M \\
        SUT base & 66M & 29.2 & 787M \\
        \midrule
        Transformer big$^f$ & 210M & 29.3 & 2090M \\
        LightConv$^g$ & 202M & 28.9 & 1750M\footnotemark[2] \\
        DynamicConv$^g$ & 213M & 29.7 & 1790M\footnotemark[2] \\
        Admin 18L-18L$^h$ & 151M & 29.0 & 1490M \\
        Admin 60L-12L$^i$ & 256M & 30.1 & 2550M \\
        MoA big$^d$ & 200M & 29.4 & 1220M \\ 
        \midrule
        UT big + SB halting & 105M & 29.6 &  3707M \\
        SUT big & 110M & 29.4 & 787M \\
        \bottomrule
    \end{tabular}
    \label{tab:bleu}

\end{table}

\section{Experiments}
First, we show that we can scale the UT with SUT on the WMT'14 English-German \citep{bojar2014findings} translation task.
We then ran experiments on Compositional Freebase Questions (CFQ; \citealt{keysers2019measuring}) to test for compositional generalization properties.
To further analyze the behaviour of the model under compositional generalization settings, we test our model on the Logical inference task from \cite{bowman2015tree}.
All experiments were implemented within the Fairseq framework \citep{ott2019fairseq}.

\subsection{English to German Translation}

\begin{table}[t]
        \caption{Ablation Study. ``-- MIM loss'' means replacing the MIM loss with the load balancing loss used in \citep{fedus2021switch}.
        ``-- MoMHA'' means replacing MoMHA with the MoA introduced in \citep{zhang2022mixture}.
        }
        \vskip -0.1in
        \centering
        \small
        \begin{tabular}{lcccc}
        \toprule
            Model & Valid loss & BLEU & \#Params \\
            \midrule
            SUT base & 2.192 & 29.2 & 66M \\
            -- MIM loss & 2.221 & 28.9 & 66M \\
            -- MoMHA & 2.232 & 28.7 & 66M \\
            -- ACT loss & 2.217 & 29.0 & 66M \\
            -- halting & 2.219 & 29.1 & 65M \\
            \bottomrule
        \end{tabular}
        \label{tab:ablation}

\end{table}

\begin{table}[t]
        \caption{FFD Expert-Word co-occurrences.}
        \vskip -0.1in
        \centering
        \small
        \begin{tabular}{lcccc}
        \toprule
            Exp. & 6 & 17 & 41 & 46 \\
            \midrule
         \parbox[t]{2mm}{\multirow{5}{*}{\rotatebox[origin=c]{90}{Top 5}}} 
             & a     & he   & ed  & team \\
             & their & they & ing & children \\
             & his   & his  & ted & police \\
             & this  & He   & y   & devices \\
             & an    & you  & red & system \\
        \midrule
        & Det. & Pronouns &  Suffixes & Nouns \\
        \bottomrule
        \end{tabular}
        \label{tab:word_occurence}
\end{table}

We perform experiments on the WMT'14 English-German translation dataset \citep{bojar2014findings}.
We use the pre-processing from  \citet{liu2020very}. 
We use a joined dictionary and share all word embeddings of the encoder and decoder. 
For evaluation, we average the last 5 best models  according to their negative log-likelihood scores. 
We report the BLEU scores \citep{papineni2002bleu}, and also report the MACs (Multiply-Accumulate Operations) to evaluate the runtime computational costs of the different models. 
MACs of previous models were computed in \citet{zhang2022mixture}.
\footnotetext[1]{The open-source tool \textsc{ptflops} (\url{https://github.com/sovrasov/flops-counter.pytorch}) is used to calculate the MACs.}
\footnotetext[2]{The MACs values of DynamicConv and LightConv are underestimated.
Because the \textsc{ptflops} does not support the customized convolution layers. }

The results are reported in Table \ref{tab:bleu}.
We compare against strong baselines while accounting for the number of parameters in these models.
In addition, we train two UTs by setting $E=1, k=1$, and parameterizing the FFD and Attention layers with parameters to match our $\sim$65M, and $\sim$110M setting for SUT.
The SUTs and UTs both demonstrate good parameter efficiency when compared to previous models.
In the $\sim$110M parameter class, SUT and UT perform at around 29.4 and 29.6 BLEU respectively, while previous models require $\sim$200M parameters.
While the SUT does not perform as well as the UT, but the computations required during runtime could be as low as one-fifth of UT.
Also, because we keep $k$ constant for SUT, the MACs stays constant as SUT scales up.

\begin{table*}[t]
    \caption{CFQ Results. Results on UT are an average of 5 runs on different seeds.}
    \centering
    \small
    \begin{tabular}{rcccccr}
    \toprule
        Model &  Pretraining & MCD1 & MCD2 & MCD3 & Avg. & MACs\footnotemark[1] \\
        \midrule
        T5-based UT ~\citep{bergen2021systematic}    & {\footnotesize \XSolidBrush} & 42.7 & 9.5  & 11.6 & 21.3 & 1154M \\
        Edge Transformer~\citep{bergen2021systematic}       & \XSolidBrush & 47.7 & 13.1 & 13.2 & 24.7 & 6504M \\
        Transformer~\citep{keysers2019measuring}          & \XSolidBrush & 42.5 & 11.2 & 10.6 & 21.4 & 1154M  \\
        \midrule
        T5~\citep{furrer2020compositional}         & \Checkmark & 61.6 & 31.3 & 33.3 & 42.1 & 1154M  \\
       
        Roberta~\citep{zheng2021disentangled}    & \Checkmark & 60.6 & 33.6 & 36.0 & 43.4  &  1660M \\  
        Dangle~\citep{zheng2021disentangled}     & \Checkmark & 78.3 & 59.5 & 60.4 & 66.1  & 51033M \\
        \midrule
        T5-based UT (ours) & \XSolidBrush & 68.3 $\pm$ 2.9 & 43.1 $\pm$ 1.5 & 45.7 $\pm$ 1.8 & 52.3 $\pm$ 1.6 & 441M \\
        UT w/o halting    &  \XSolidBrush & 71.0 $\pm$ 3.5 & 48.6 $\pm$ 2.3 & 51.3 $\pm$ 0.2 & 56.9 $\pm$ 1.5 & 654M \\
        UT with halting   &  \XSolidBrush & 72.4 $\pm$ 3.5 & 51.1 $\pm$ 1.8 & 51.7 $\pm$ 2.3 & 58.4 $\pm$ 1.2 & 654M \\
    \bottomrule
    \end{tabular}
    \label{tab:cfq}
\end{table*}

We ran experiments removing different aspects of the model and its training process, including: MIM auxiliary loss, Mixture of MHA, the ACT loss, and the halting mechanism.
The results are in Table \ref{tab:ablation}.
The introduction of multiple heads to the MoA was crucial in seeing performance gains on this task, as well as having the MIM loss as a load-balancing auxiliary objective.
Interestingly, halting does contribute as much of a performance gain as it does in CFQ.

Additionally, we compute the top 5 tokens that occur in conjunction with each expert, regardless of layers, and find that certain associations exist.
We pick several experts in Table \ref{tab:word_occurence} that show a clear sign of co-occurring with tokens that seem to show a pattern.
This suggests that while there may be redundancy between the experts, groups of experts can specialize on certain tasks, resulting in some modularity.
Future work can investigate if such modularity can result in more robust generalization.

\subsection{Compositional Freebase Questions}

We run experiments on the Compositional Freebase Questions (CFQ; \citealt{keysers2019measuring}) dataset to determine the compositional generalization abilities of the SUT.
This is a translation task from natural language to a SPARQL query.
As an example, the sequence \texttt{Who wrote M1 and wrote a film} would be translated to the target sequence 
\texttt{SELECT DISTINCT ?x0 WHERE \{ ?x0 a people.person . ?x0 film.writer.film ?x1 M1 . ?x1 a film.film \}}.
CFQ tests for compositional generalization using the notion of \emph{compound divergence}, which measures how different the training set and test set are in terms of combinations of tokens, which they refer to as compounds.
To our knowledge, the current best-performing models either finetune a pretrained language model or, use knowledge about the task to design a suitable prompt for a large language model \citep{drozdov2022compositional}.
While the prompting approach is extremely effective at the CFQ task, we view the task as a benchmark for compositional generalization in general and should be viewed in concert with other experiments, especially real-world data (like translation).
When using domain knowledge of the task in the prompt, the results may indicate better performance with a specific approach for CFQ (and perhaps other SQL translation tasks) but might be difficult to extrapolate to other settings.

In our experiments, we use preprocessing scripts from \citet{zheng2021disentangled}.
The scripts perform preprocessing to the target sequence that simplifies the target sequence the same way performed in \citet{furrer2020compositional}.
Accordingly, we train a baseline Transformer on the transformed target.
We performed a search on the SUT hyperparameters, using the MCD1 validation set,  and the best-performing set of parameters are 
Attention $(E=1, k=1, H=8, D=64, W = 1)$ and FFD $(E=1, k=1, D=1024)$, which corresponds to the UT setting. 
Refer to Appendix \ref{app:training} for further details.
Since CFQ is a relatively small task, larger scale is not a factor and might suggest that expert specialization may not be as helpful.
The results are shown in Table \ref{tab:cfq}.
In cases with and without halting, the model already outperforms previous benchmarks, including the UT baseline from \citet{bergen2021systematic}.
For a fairer comparison, we use the same hyperparameters as our UT implementation, we modify our UT implementation to be more similar to the T5-based UT in \citet{bergen2021systematic}.
These changes include: the bucketed relative position bias used by T5, and going from post layer-norm to pre layer-norm.
While this results in  much improved results compared to the original paper, our implementation of UT still outperforms it.

The Dangle \citep{zheng2021disentangled}  model, which beats our model, also requires re-running the encoder for every token decoded.
This is an expensive process, but given that both our method and Dangle perform well at this task, is additional evidence that iterative processes are beneficial for compositional generalization.

\subsection{Logical Inference}
\begin{table}[t]
\small
    \centering
    \small
    \setlength{\tabcolsep}{4.5pt} 
    \caption{
    Test accuracy of the models, trained on operation lengths of $\leq 6$, with their out-of-distribution results shown here (lengths 7-12).
    LSTM baseline from \citet{bowman2015tree}, and Transformer baseline from \citet{shen2019ordered}}
    \begin{tabular}{l cccccc ccc }
    \toprule
    \textbf{Model} & \multicolumn{6}{c}{\textbf{Number of Operations}} & \multicolumn{3}{c}{\textbf{Comp. Gen.}}\\
          & 7 & 8 & 9 & 10 & 11 & 12 & A & B & C  \\
    \midrule
    LSTM     & 88 & 84 & 80 & 78 & 71 & 69  & 80 & 60 & \textbf{59}  \\
    Transformer & 51 & 52 & 51 & 51 & 51 & 48 & 53 & 51 & 51 \\
    \midrule
    SUT &   \textbf{98} & \textbf{97} & \textbf{94} & \textbf{90} & \textbf{88} & \textbf{81} & \textbf{97} & \textbf{94} & 52 \\
    \bottomrule
    \end{tabular}
    \label{tab:proplogparse}

\end{table}

We use the logical inference task from \cite{bowman2015tree} as a test bench for UT.
Despite the apparent simplicity of the language, the task inherently requires the model to learn  the hierarchical structure of the problem. 
Each instance of the task comprises of two logical statements, and the goal is to predict if the statements are equivalent, contradictory, disjoint, or entail in either direction.
For example, given $s_1 =$ \texttt{a} and $s_2 =$ \texttt{a ( or b )},  then $s_1 \sqsubset s_2$.
The crux of the task is in training the model on sequences that have 0-6 logical operators and evaluating it on sequences that have 7-12 operators.
Given our sequence-to-sequence setting, we convert the task into a translation task.
The model takes \texttt{sentence1 \#SEP\# sentence2} as its source sentence, with the target sentence being the single-token label for that pair.

We train a 12 layer model with Attention $(E=12,k=4, H=2, D=32, W=1)$ and FFD $(E=12, K=4, D=128)$ and halting. Refer to Appendix  \ref{app:training}  for further details.
Training a 12-layer Vanilla Transformer achieves approximately the same results as in \citet{shen2019ordered}, so we report their results.
Our results in Table \ref{tab:proplogparse} confirm the findings of \citet{tran2018importance}, showing that with recurrence in SUTs, we are able to generalize to longer sequences of the task.
While there are other models that induce a tree structure that performs exceedingly well on the task, we wanted to evaluate our model against other popular architectures.
The LSTM is a strong baseline, and we find that UT outperforms it in generalization.
We also evaluate UTs on the compositional generalization splits as proposed in \cite{shen2019ordered}, where the splits A, B, and C are in increasing difficulty.
The results show that UTs are able to generalize better for the A and B splits, outperforming the LSTM and VT. 
Split C is still presents a challenge for the Transformer variants.

\begin{figure}[t]
\begin{center}
\includegraphics[width=\linewidth]{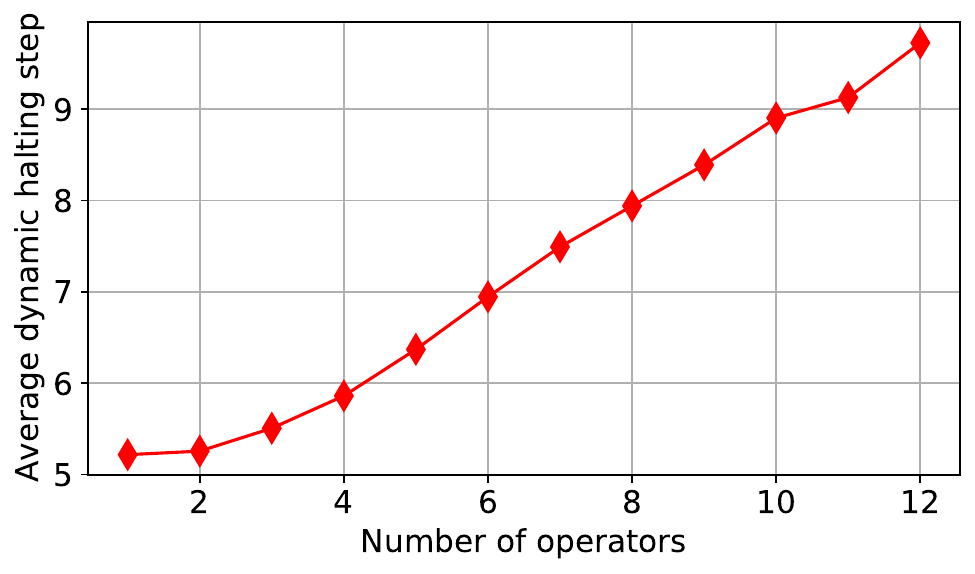}

\caption{
The average dynamic halting depth of the UT model as the number of operators increases in the test set. 
The model learns to think more when the problem is harder.
}
\vskip -0.2em
\label{fig:depthincreases}
\end{center}
\end{figure}

Additionally, we compute the average halting depth for the test data segmented by operator counts.
Because more operators require more nesting of expressions in the sequences, more recursion is required to properly parse the sequence.
As expected, in Figure \ref{fig:depthincreases}, the average halting depth increases as more operators are used.
The operator count for these clauses are correlated with length, which suggests that SUTs may be suited to generalize for length.
We include further experiments on length generalization in the Appendix Table \ref{tab:lralistops}.

\subsection{Post-training Computation Reduction}
Does lowering $\alpha_\text{thresh}$ \emph{after} training cause the model to halt earlier, saving computation? How much would that cost us in terms of accuracy?

We estimate the skipped SUT block computations given different values of $\alpha_\text{thresh} \in \{0.1, 0.2, \hdots, 0.9\}$ by looking at the halting patterns of the decoder given the ground truth source-target pairs. 
We pass the source-target pair into the model and analyze the halting patterns of the model, giving us a rough estimate of how much computation would be saved as a percentage of computing all layers of the SUT.

\paragraph{Logical Inference}
\begin{figure}
    \centering
    \includegraphics[width=0.9\linewidth]{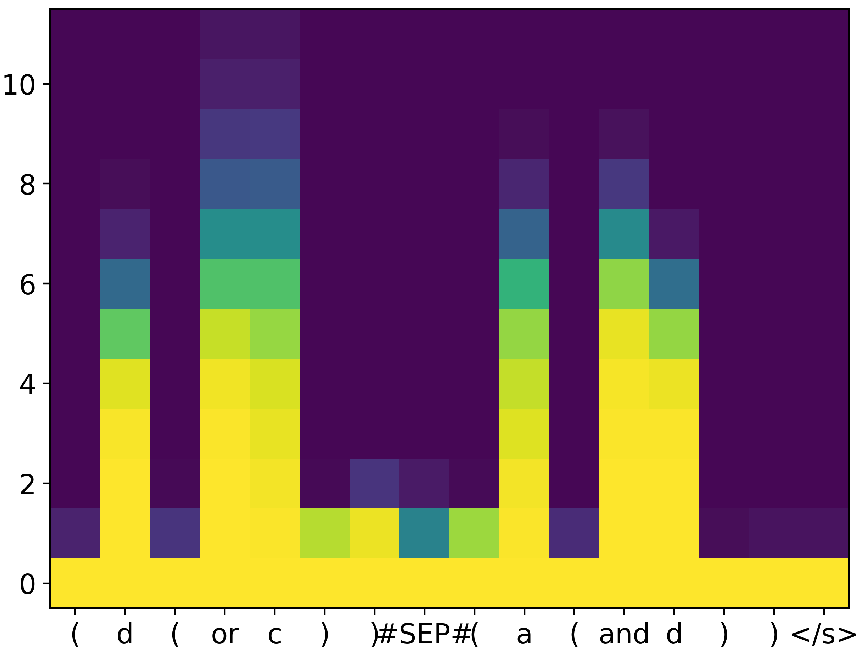}   \\
    \includegraphics[width=\linewidth]{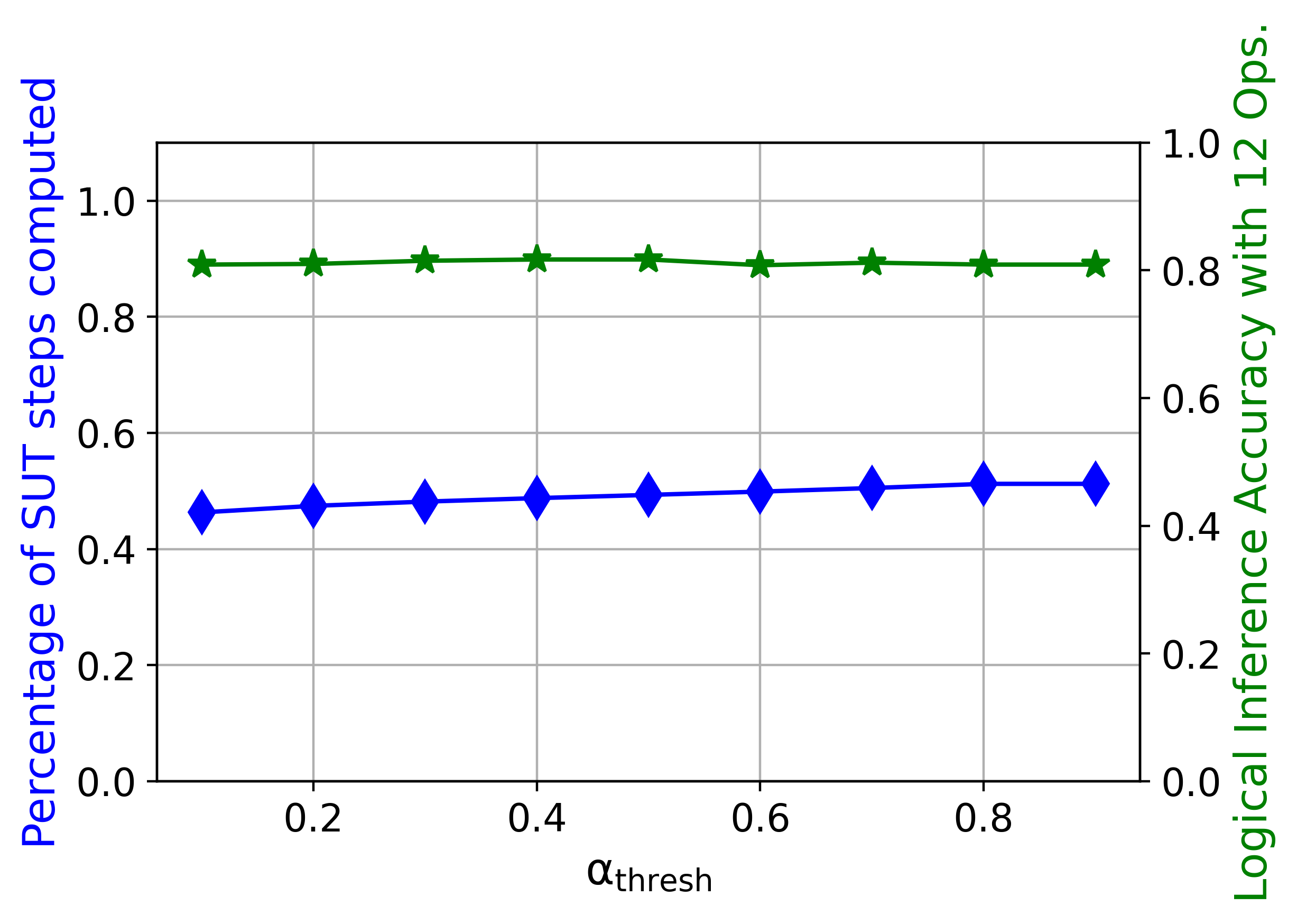}
    \caption{\emph{Above:} Plot of $1 - \sum_{l'=1}^{l-1}  \alpha^{(t)}_{l'}$, for an example Logical Inference input --- $x$-axis: timesteps, $y$-axis: layers.
    This visualizes the halting pattern of the model:  dark blue represents halted, while yellow represents active.  \emph{Below:} Efficiency vs. Performance tradeoff curves when $\alpha_\text{thresh}$ is adjusted.}
    \label{fig:halting_proplog}
    \vspace{-1em}
\end{figure}
We observe the resulting performance on the hardest split of the test set with 12 operations.
Due to the already saturated halting pattern, the halting probability $\alpha_l$ spikes rapidly from close to 0 to higher values, 
resulting in a near constant $\sim$ 50\% reduction of the computation time regardless of the threshold.

\paragraph{CFQ}
\begin{figure}
\vspace{-0.1em}
    \centering
    \includegraphics[width=\linewidth]{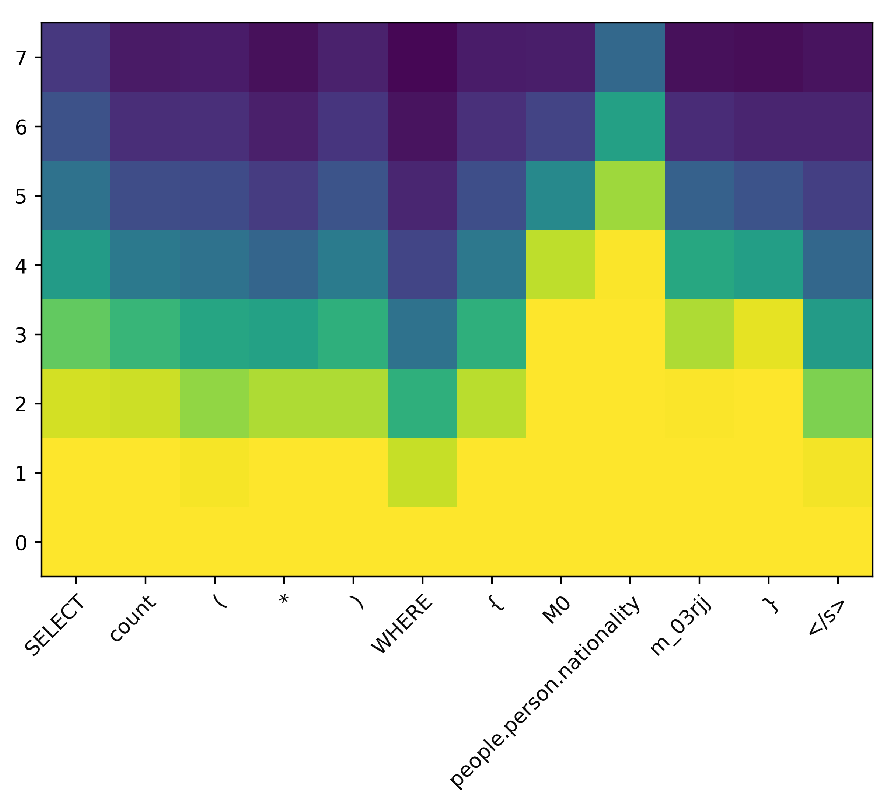}    \\
    \vspace{-1em}
    \includegraphics[width=\linewidth]{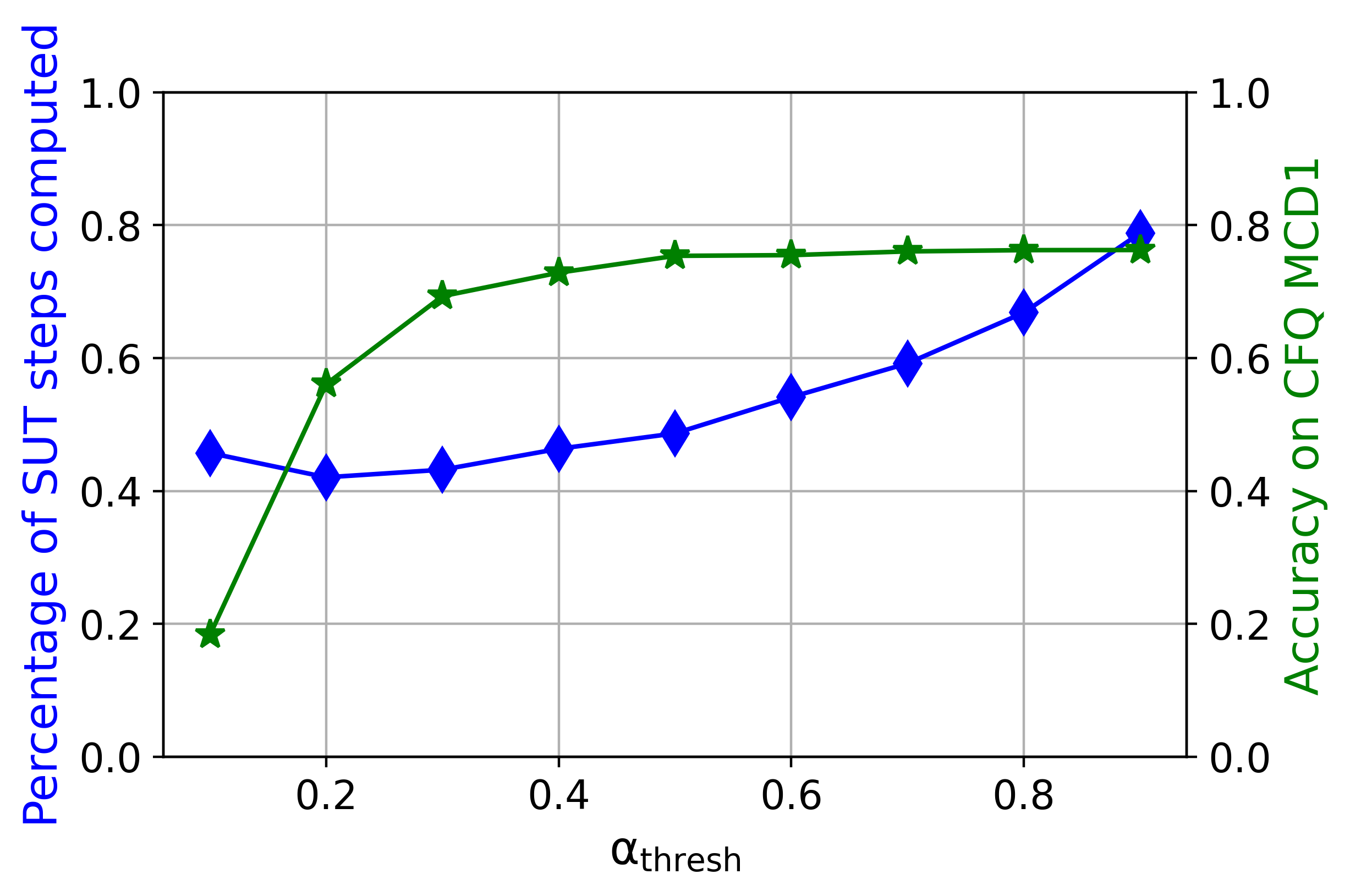}
    \caption{Halting plot and trade-off curves for CFQ. (See Figure \ref{fig:halting_proplog} for description)}
    \label{fig:halting_cfq}
\end{figure}
Using the MCD1 test split of the dataset, and our best-performing model on MCD1, we perform the $\alpha_\text{thresh}$ adjustment.
The halting patterns reflect the repeated structure of SQL, using fewer steps for `.` and `WHERE`, while the main bulk of the region within \texttt{\{...\}} requires more SUT steps before halting.
Surprisingly, when $0.8 \leq \alpha_\mathrm{thresh} \leq 0.999$, the accuracy remains fairly constant.
An estimated 33\% of the computation steps were skipped at $\alpha_\mathrm{thresh} = 0.8$. 
At $\alpha_\text{thresh}=0.1$, there is a slight increase in the number of computed steps, which is possible since halting earlier will  result in different embeddings, and result in different halting decisions in other timesteps.
Overall, the results suggest that we can save about 20\% of the SUT computation steps without any drop in accuracy, and about $\sim$50\% for a 0.2\% decrease.

\paragraph{English-German Translation} 

\begin{figure}
\vspace{-2em}
    \centering
    \includegraphics[width=\linewidth]{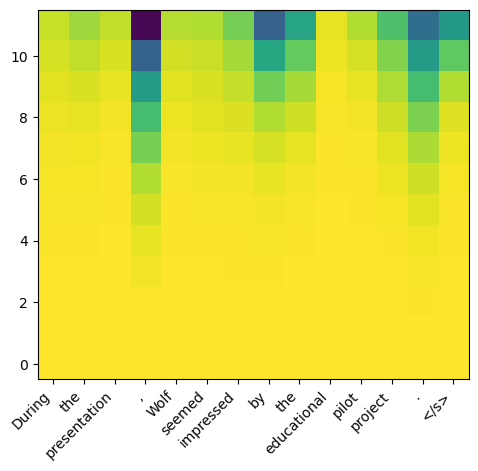}    \\
    \vspace{-1em}
    \includegraphics[width=\linewidth]{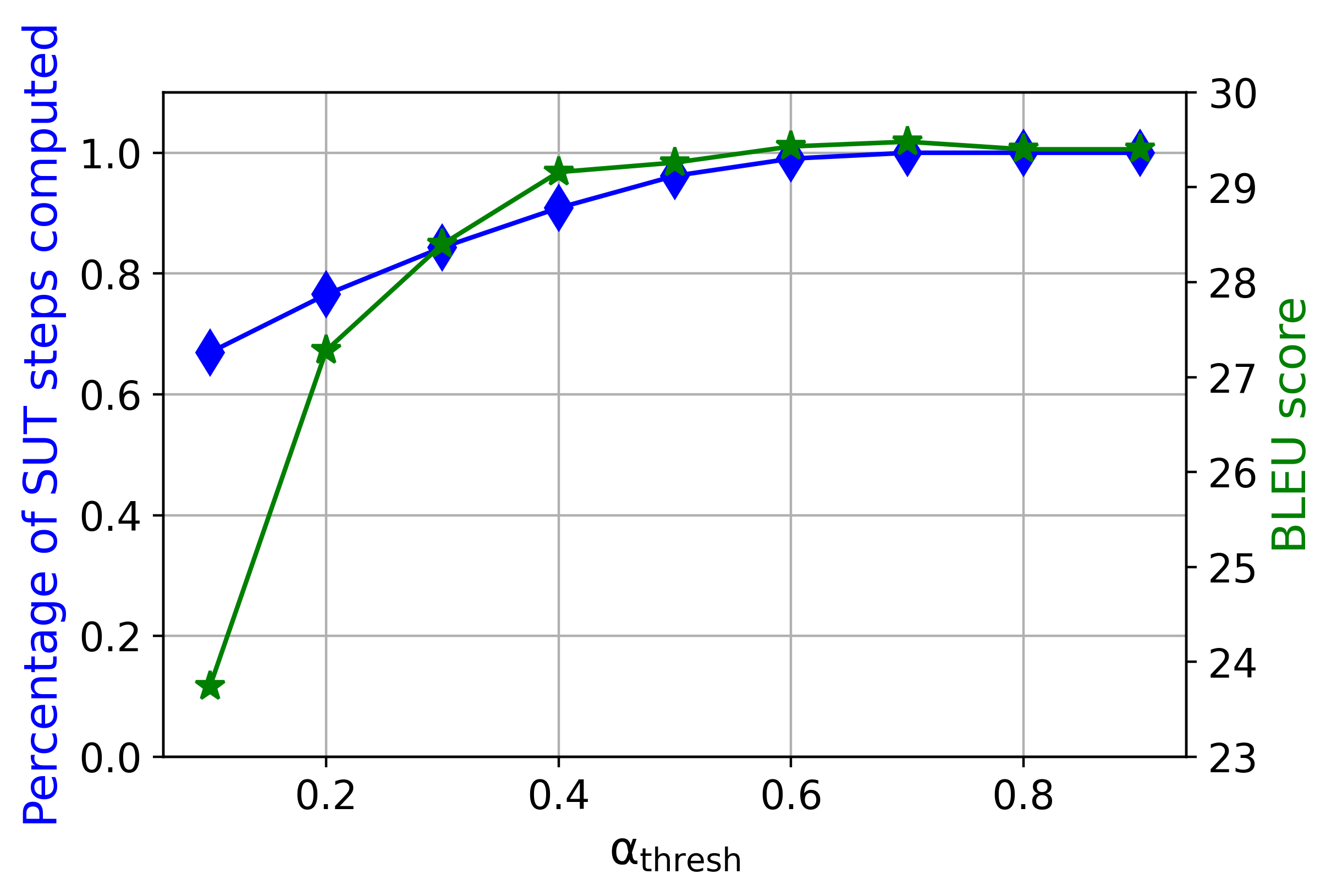}
    \caption{Halting plot and trade-off curves for English-German Translation. (See Figure \ref{fig:halting_proplog} for description)}
    \label{fig:enter-label}
\end{figure}
For this larger dataset, we find that these translation models halt much later, suggesting that the translation task requires more computational steps than the 6-layer SUT we used.
However, further increasing the number of layers to 12 layers does not bring about much benefit, as evidenced by the halting in Figure \ref{fig:depthincreases}, which is an example of the halting mechanism using nearly all layers.
For comparison, Admin 60L-12L model, requires a 60-layer encoder to achieve its performance.
Even when $\alpha_\text{thresh} = 1$, the skipped computation steps remain at about 33\%, compared to 80\% in the CFQ task.
We find that we can reduce the computation by 9\% while still retaining a BLEU score of 29.1.

\section{Conclusion}
We show that it is possible to scale up the UT via SUTs, and SUTs outperforms models of the same capacity in the WMT'14 English-to-German translation task.
The recursive nature of both UTs and SUTs allows for better inductive biases, which we have demonstrated in synthetic tasks like CFQ and logical inference.
VTs have been shown to be poor at these compositional generalization tasks without additional domain knowledge.
The stick-breaking dynamic halting mechanism also allows post-training adjustment of computation cost, which is a boon for deployment at scale.
\paragraph{Limitations}
While the experiments in this paper show the desirable generalization properties of UTs, there are some aspects of compositional generalization that SUTs do not solve.
Importantly, while we demonstrate scaling UTs up via SMoEs, further experiments on larger settings are needed to ascertain viability in large scale systems.
Other issues may also crop up in the further scaling of SUTs, but we believe there is ample literature to draw on to finding solutions for these problems.

\bibliography{anthology,custom}
\bibliographystyle{acl_natbib}
\clearpage
\appendix

\section{Experimental Details}
\label{app:training}

\begin{table*}[t]
    \centering
    \small
        \caption{Hyperparameters for different models. 
    }
    \begin{tabular}{llcccccccc}
        \toprule
        Dataset & Model & $N_{att}$ & $N_{ffd}$ & Emb Size & $D_{att}$ & $D_{ffd}$ & \#Layers & \#Params\\ \midrule
        Logical Inference & SUT & 12 & 12 & 512 & 64 & 128 & 12-12 & 7.6M  \\
        \midrule
        CFQ & SUT (with / without halting)  & 1 & 1 & 512 & 512 & 1024 & 8-8 & 7.2M / 6.7M \\
        \midrule
        \multirow{4}{*}{WMT14 En-De} 
        & UT base & 1 & 1 & 960 & 960 & 3840 & 6-6 & 64M \\
        & UT big & 1 & 1 & 1344 & 1344 & 5376 & 6-6 & 105M \\
        & SUT base & 24 & 24 & 512 & 256 & 512 & 6-6 & 66M \\
        & SUT big & 48 & 48 & 512 & 256 & 512 & 6-6 & 110M \\

        \bottomrule
    \end{tabular}

    \label{tab:model_hyperparameters}
\end{table*}

\begin{table*}[t]
    \centering
    \small
        \caption{Training Hyperparameters for different models. 
    The BSZ represent the maximum number of tokens in each batch.
    }
    \begin{tabular}{lcccccccc}
        \toprule
        Dataset & BSZ & LR & warmup & Dropout & DropATT & DropFFD & DropGAT & Epochs \\ \midrule
        Logical inference & $16384 \times 4$ & 7e-4 & 4000 & 0.5 & 0.2 & 0.5 & 0.1 & 450\\
        CFQ & $16384 \times 4$ & 7e-4 & 4000 & 0.5 & 0.2 & 0.5 & 0.1 & 450\\
        WMT14 EN-DE & 8192 $\times$ 32 & 7e-4 & 4000 & 0.2 & 0.2 & 0.1 & 0.2 & 100 \\
        \bottomrule
    \end{tabular}

    \label{tab:hyperparameters}
\end{table*}
All of our models are trained on V100 GPUs.
We use the Adam Optimizer with $\beta_1= 0.9$, $\beta_2= 0.98$ and $\epsilon = 1e-9$. 
We use a inverse square root learning rate scheduler.
During training, we employed label smoothing~\citep{szegedy2016rethinking} of value 0.1. 
More training hyperparameters can be found in Table~\ref{tab:hyperparameters}.

\section{Additional Experiment Results}
\subsection{Logical Inference} \label{app:proplog}
\paragraph{Expert specialization} we can inspect the routing frequencies to each of the experts with respect to the input token.  
We compute the pointwise mutual information (PMI; \citealt{church1990word}) of each word to each expert at all layers.
Looking at the MoMHA experts, we see some patterns that suggest the experts have specialized in their tasks.
For example, the \texttt{or} operator is dominating expert 9.
Interestingly, the \texttt{not} operator is reliant on experts 4 and 5, unlike the \texttt{or} and \texttt{and} operators, which are binary and rely on experts 1 and 2. 
The variables \texttt{a} through \texttt{f} seem to have similar expert usage patterns as well.
This suggests that the experts are specialized in their operation.

\begin{figure}[H]
\begin{center}
\includegraphics[width=0.8\linewidth]{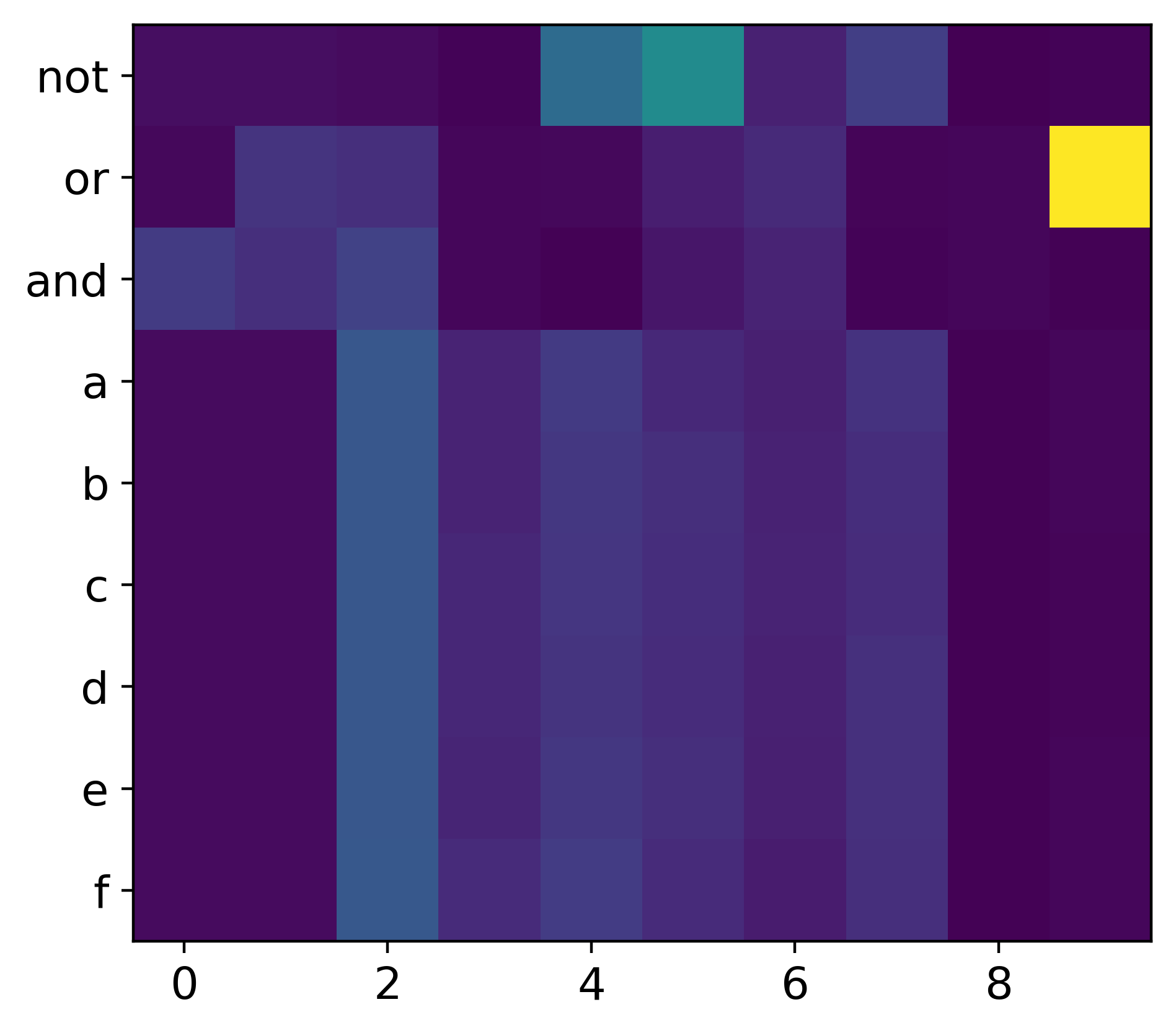}
\caption{Expert-token affinity for the logical inference task. We plot the PMI scores between the tokens ($y$-axis) and the experts ($x$-axis). 
Looking at the MoMHA experts, we see some patterns that suggest the experts have specialized in their tasks.
For example, the \texttt{or} operator is dominating expert 9.
Interestingly, the \texttt{not} operator is reliant on experts 4 and 5, unlike the \texttt{or} and \texttt{and} operators, which are binary and rely on experts 1 and 2. 
The variables \texttt{a} through \texttt{f} seem to have similar expert usage patterns as well.
This suggests that the experts are specialized in their operation.
}

\label{fig:expert_operator_affinity}
\end{center}
\end{figure}

\subsection{Long Range Arena Listops}
\begin{table}[H]
        \caption{Results on Long-range Arena ListOps benchmark. We include the Transformer and top 2 results from \citet{tay2020long} for brevity, and the best performing external result on the leaderboard at \url{https://github.com/google-research/long-range-arena}}
        \centering
        \begin{tabular}{lc}
        \toprule
            Model & Accuracy \\
            \midrule
            Transformer & 36.37 \\
            Reformer    & 37.27 \\
            Synthesizer & 36.99 \\
            IGLOO \citep {sourkov2018igloo}   & 39.23 \\
            \midrule 
            SUT & \textbf{49.23} \\
            \bottomrule
        \end{tabular}
        \label{tab:lralistops}
        \vskip -0.1in
\end{table}

\subsection{SCAN \& PCFG Generalization}
In all cases, we ran a baseline Vanilla Transformer (VT) model using the IWSLT Transformer model from fairseq, and compared with a relevant VT baseline from the literature.
We were unable to replicate the \texttt{add\_turn\_left} result in \citet{furrer2020compositional} on SCAN.
We also found that our VT results on the PCFG splits were slightly higher.

These results show that there are nuances to compositional generalisation.
While the SUT can bring consistent benefits in the case of PCFG, the harder generalisation aspects of \texttt{add\_jump} and length in SCAN may require a different kind of generalisation. To our knowledge, there is work that deals explicitly with the \texttt{add\_jump} \citep{russin2019compositional, li2019compositional} and others that deal with the length split by decoding in a tree structure \citep{tan2020recursive,kim2021sequence}.
SUT tackles some of these aspects of compositional generalisation, but not all, and this further illustrates the generalization SUTs can aid in.

\begin{table}[H]
\caption{Experimental results for the Vanilla Transformers (VT) and a Universal Transformer (SUT with 1 expert). results are full-string match accuracy.  $^1$\citet{furrer2020compositional} $^2$\citet{hupkes2020compositionality}} 
    \centering
    \small
    \begin{tabular}{llccc}
    \toprule
        \multicolumn{2}{l}{Task}  & VT  & VT (ours)  & UT \\
    \midrule
     \parbox[t]{2mm}{\multirow{3}{*}{\rotatebox[origin=c]{90}{SCAN}}} 
        & \texttt{add\_turn\_left} & 100$^1$  & 20  & 100 \\
        & \texttt{add\_jump}       & 1$^1$  & 7  & 0 \\ 
        & \texttt{length}          & 0$^1$  & 0  & 7 \\ 
    \cmidrule{1-5}
    \parbox[t]{2mm}{\multirow{2}{*}{\rotatebox[origin=c]{90}{PCFG}}} 
        & \texttt{systematicity}   & 72$^2$  & 76  & 87 \\ 
        & \texttt{productivity}    & 52$^2$  & 56  & 62 \vspace{1mm} \\
    \bottomrule
    \end{tabular}
\end{table}~

\subsection{WMT'14 English to German}\label{app:ende_translation}

\begin{figure}[H]
\begin{center}
\includegraphics[width=.9\linewidth]{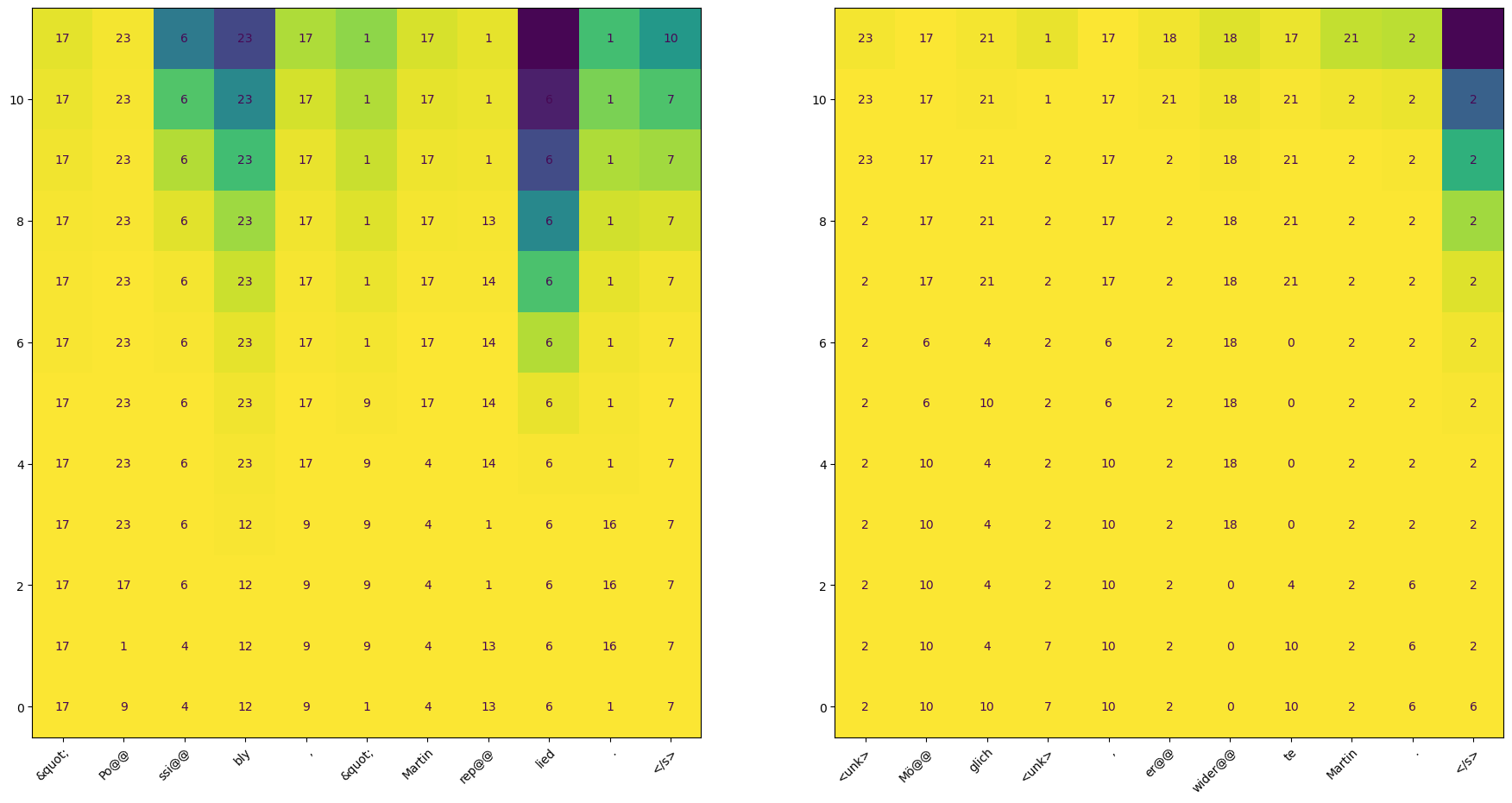}
\caption{
Halting plot for a 12-layer SUT encoder and decoder. Number represents the top expert being used.
For WMT'14 English-German, we find that these translation models halt much later, suggesting that the translation task requires more computational steps than the 6-layer SUT we used.
However, further increasing the number of layers to 12 layers does not bring about much benefit, as evidenced by the halting in Figure \ref{fig:depthincreases}, which is an example of the halting mechanism using nearly all layers.
Looking  at the top-scoring expert for the plot, it appears the experts selected are largely word dependent, and tend to repeat. 
}
\label{fig:haltingplot}
\end{center}
\end{figure}

\end{document}